\documentclass[a4paper]{article}
\usepackage{Odyssey2020}
\usepackage{epsfig,amssymb,amsmath}
\usepackage{multirow}
\usepackage{hyperref}
\usepackage{xcolor,colortbl}
\definecolor{LightCyan}{rgb}{0.88,1,1}
\definecolor{Gray}{gray}{0.85}

\ninept

\title{Comparison of Speech Representations for Automatic Quality Estimation in Multi-Speaker Text-to-Speech Synthesis}
\makeatletter
\def\name#1{\gdef\@name{#1\\}}
\makeatother
\name{{\em Jennifer Williams, Joanna Rownicka, Pilar Oplustil, and Simon King}}
\address{The Centre for Speech Technology Research, University of Edinburgh, United Kingdom\\
{\small \tt {j.williams@ed.ac.uk} } }
\begin{document}
\maketitle

\begin{abstract}
We aim to characterize how different speakers contribute to the perceived output quality of multi-speaker Text-to-Speech (TTS) synthesis. We automatically rate the quality of TTS using a neural network (NN) trained on human mean opinion score (MOS) ratings. First, we train and evaluate our NN model on 13 different TTS and voice conversion (VC) systems from the ASVSpoof 2019 Logical Access (LA) Dataset. Since it is not known how best to represent speech for this task, we compare 8 different representations alongside MOSNet frame-based features. Our representations include image-based spectrogram features and \textit{x}-vector embeddings that explicitly model different types of noise such as T60 reverberation time. Our NN predicts MOS with a high correlation to human judgments. We report prediction correlation and error. A key finding is the quality achieved for certain speakers seems consistent, regardless of the TTS or VC system. It is widely accepted that some speakers give higher quality than others for building a TTS system: our method provides an automatic way to identify such speakers. Finally, to see if our quality prediction models generalize, we predict quality scores for synthetic speech using a separate multi-speaker TTS system that was trained on LibriTTS data, and conduct our own MOS listening test to compare human ratings with our NN predictions. 
\end{abstract}

\section{Introduction}
The ability to automatically judge the quality of speech is an open and active research area, gaining more attention recently due to the predictive power of deep neural networks (NNs). Speech quality assessment can be used in various wide-ranging tasks, from speech enhancement to voice conversion (VC) and text-to-speech (TTS) synthesis. In particular, the ability to automatically assess synthetic speech would result in large gains in the field such as faster and more efficient evaluation savings from expensive listening tests. If there was an automatic speech quality prediction tool, it could be used for rapid assessment and comparison of models and it would facilitate design decisions including stopping criterion for TTS training.

It is widely accepted that some speakers give higher quality than others for building a TTS system. This is true across different types of TTS systems, from HMM-based and DNN-based statistical parametric speech synthesis (SPSS), to encoder-decoder models and neural vocoders. Likewise, it is becoming more commonplace to use large multi-speaker corpora for TTS training~\cite{gibiansky2017deep}. Often these large corpora of natural speech, such as LibriTTS~\cite{zen2019libritts}, are of variable recording quality. While it was once preferable to train TTS systems on speech from studio-quality professional voice actors - the multi-speaker corpora are very useful for multi-speaker TTS. Multi-speaker TTS systems, such as ~\cite{gibiansky2017deep}, learn simultaneously a common sound-to-phoneme mapping and the nuances of each speaker's individual voice. 

One recent advancement in automatic speech quality assessment was MOSNet~\cite{lo_mosnet:_2019}, which extended similar work from Quality-Net~\cite{fu_quality-net:_2018}. The MOSNet system was trained to predict mean opinion scores (MOS) as a measure of naturalness for voice-conversion, based on human judgments of MOS as the ground truth. The MOSNet system is free and open-source, and provides three different pre-trained models. Since voice conversion and TTS tasks are somewhat related, we were hopeful to apply the pre-trained models from MOSNet to our TTS system and data. However, in our preliminary experiments we found that the pre-trained VC models did not generalize well.  

Automatic quality assessment is inherently very difficult. It requires a large dataset that has multiple speakers, and multiple systems, and must be consistently labeled with quality scores. As a machine learning task, previous work has shown that regression models will tend to learn an average score and that it is more difficult for the NN to predict high and low outliers ~\cite{lo_mosnet:_2019}. A related problem comes from human judgments of quality. If human listeners are not exposed to a wide variety of poor and high quality speech during the listening test, there is a risk that many utterances will be marked ``average''. It is also widely accepted that MOS scores do not reflect the complex nature of a quality spectrum since the relationship between values and quality is not linear (i.e. the difference between MOS scores of 1 and 2 $\neq$ difference between 3 and 4). There is no guarantee that a NN trained for one TTS system would generalize to another TTS system. In order to help our model generalize to new systems, we train on the ASVSpoof 2019 LA dataset because this dataset consists of 13 different systems (7 TTS, 3 TTS+VC, and 3 VC) that range in quality, and use the same speakers. More details about the LA dataset are provided in Section~\ref{sec:asv_data}. 

In this paper, we experiment with training different representations to find a MOS model that can generalize to our own multi-speaker TTS system. We build on previous work from ~\cite{lo_mosnet:_2019} with the following contributions:

\begin{enumerate}
    \item Re-train the original MOSNet framework using the LA dataset and explore frame-based weighting in the loss function
    \item Train a new low-capacity CNN architecture on LA dataset and compare 8 types of speech representations 
    \item Characterize NN prediction performance based on speaker-level rankings
\end{enumerate}

We are particularly interested in finding a way to identify speakers who make the highest and lowest quality speech. In addition to the above, we also incorporate a ranking-based evaluation metric for our models and representations. This allows us to better understand how speakers compare to each other in a given system, or across multiple systems. We seek to identify a model and representation that predicts MOS scores with the highest correlation to human judgments, and also preserves the relative quality ranking of speakers.

\section{Background}

\subsection{Estimating Synthetic Speech Quality}
Speech quality estimation is an important area of research, especially for synthetic speech, and has garnered an incredible amount of attention recently. While automatic methods are well-motivated, this problem is inherently difficult due to variability in ground truth human judgments. MOS ratings are one of many de-contextualized TTS metrics. Ideally these listening tests would convey specific application contexts to calibrate to listeners expectations. That is why \cite{wagner_speech_2019} calls for developing a set of best practices for TTS evaluation. Those best practices include a critical analysis the effectiveness of automatic tools, such as MOSNet~\cite{lo_mosnet:_2019}, QualityNet~\cite{fu_quality-net:_2018}, and AutoMOS \cite{patton_automos:_2016}. 

Speech quality can be characterized in different ways. For example, in \cite{mittag_quality_2019} they examined the perception of different types of artifacts that can occur during speech transmitted and lead to loss and degraded quality. Among these artifacts were 1) the presence of noise (circuit, background, or poor coding), 2) coloration from frequency response distortion, and 3) non-stationary distortions such as packet loss or clipping. They showed that a CNN-LSTM network could be trained to predict a MOS score for each of the three artifact types, with performance better than existing techniques using Pearson \textit{r} correlation and root mean square error (RMSE) metrics. 

\subsection{Speech Representations}
Many speech applications, including automatic quality assessment, make use of frame-based features. These are constructed using a sliding window across a speech spectrogram. Representations highlight or emphasize particular types of information. For example, some speech quality estimation may require representations that model phonemes (intelligibility) while others focus on modeling reverberation (naturalness). There is a gap in the research for sufficiently broad studies that compare representations to determine which ones are best for quality. 

The representations proposed in \cite{williams2019speech} show it is possible to train an \textit{x}-vector extraction system to model different spoofing environments such as room size, reverberation time, and device quality. In that case, the embedding was trained to optimize detection of different artifacts and noise. Embeddings were obtained by passing an utterance forward through the trained \textit{x}-vector system. They showed these representations to be useful for detecting speech intercept spoofing attacks. 

Speech-based CNN embeddings were proposed in \cite{rownicka_analyzing_2018} for the purpose of acoustic model adaptation. They showed that a deep CNN model trained on filter-bank features could learn information about speaker, gender, and channel noise. While they demonstrated CNN representations worked well for the original case of acoustic modeling, they also show that these embeddings could be applied to differentiate between acoustic conditions, so this representation is also of interest in our work. 

In terms of image-based representations, the Deep Spectrum (DS) features are also generated from a deep CNN, however in this case the model has been trained on a very large image dataset rather than speech data \cite{cummins_image-based_2017}. In this case a spectrogram is given as input, and the representation is provided as a vector containing activations from a particular layer in the network. These have proven useful for speech emotion recognition as well as analysis involving snoring noises \cite{amiriparian_snore_2017}.

\section{Data}
We used different datasets for different purposes and pre-training. We used the ASVSpoof 2019 Logical Access Dataset for training our MOS prediction network. We used the ASVSpoof 2019 Physical Access Dataset to train our \textit{x}-vector embeddings, which is described in more details in Section~\ref{sec:xvector_descr}. Finally, we used the LibriTTS Dataset when training our own TTS system which is described in more details in Section~\ref{sec:ophelia_description}. 

\subsection{ASVSpoof 2019 Dataset}
\label{sec:asv_data}

The ASVSpoof 2019 Dataset\footnote{https://datashare.is.ed.ac.uk/handle/10283/3336} was designed for the Third Automatic Speaker Verification Spoofing and Countermeasures Challenge~\cite{spoofing2019}. The goal of the challenge was to discriminate between genuine and spoofed speech samples. Two main types of attacks were considered in the challenge: Logical Access (LA) and Physical Access (PA). In both tasks, the datasets are derived from VCTK corpus\footnote{https://datashare.is.ed.ac.uk/handle/10283/2651}, comprising 107 speakers (46~males, 61~females). They are partitioned into three datasets (train/dev/eval) with non-overlapping speakers, with 20, 10, and 48 speakers in the datasets, respectively.

\subsubsection{Logical Access (LA) Dataset}

The LA track of the challenge centers on attacks with the use of state-of-the-art TTS and VC systems. The details of this dataset, including the 13 different TTS and VC systems, is described in~\cite{wang2019asvspoof}. For our work, we used the evaluation subset of the LA data, wherein each utterance had been rated for MOS quality scores on a scale of 1 (definitely machine) to 10 (definitely human). For the MOS test, listeners were instructed to imagine they are working at a call center and must determine if speech from an incoming call is human or machine \cite{wang2019asvspoof}. Each utterance was rated with one MOS score, as the effort was to balance ratings between human and spoofed examples. 
In the data split we used similar proportions as ~\cite{lo_mosnet:_2019}. Each split consisted of the same systems, but different speakers. Our test set had a total of 9 speakers with approximately 320 utterances per speaker evenly balanced across the 13 TTS/VC systems.

\subsubsection{Physical Access (PA) Dataset}
The PA track of the challenge addresses the countermeasures for replay spoofing attacks. The speech samples in this dataset are simulated, to assure controlled experimental conditions. Different environments (determined by room size, reverberation time, talker-to-ASV distance) and different attacks (determined by attacker-to-talker distance, replay device quality) are the factors of variation considered in the simulation. 

\subsection{LibriTTS Dataset}\label{sec:libri}
The LibriTTS dataset ~\cite{zen2019libritts} is a subset corresponding to an improved version of the  LibriSpeech dataset, originally collected for speech recognition tasks. LibriTTS was designed and released by Google, aiming to be used for TTS applications, removing noisy data and normalizing text, among other improvements. It contains a wide range of speakers reading aloud books (e.g. audio-books). As each speaker can record themselves anywhere and with any equipment, there is a huge diversity of quality, channels and room noise through out the data. Despite the cleaning process the corpus was subject to, in our previous experiments with this data we have noticed how this variety of conditions can affect the TTS models.

\subsection{Speech Representations}
Since we were not able to find any studies that compared different types of input features or speech representations for this task, we explored 8 additional representations alongside our experiments that were used to re-train the original MOSNet. The frame-level features in the original MOSNet were extracted from a sliding window over the magnitude spectrogram of each utterance. We hypothesized that some of the artifacts which make speech sound less natural (or lower quality) might be found from image-based embeddings of the spectrogram if the representation can model very high resolution artifacts. And we hypothesized that TTS might have similar types of detectable artifacts, such as reverb, that could be modeled using \textit{x}-vectors. 

\subsubsection{Deep Spectrum (DS) Features}
DS features are created by forwarding an utterance-level speech spectrogram through a very deep CNN that was originally trained for image-recognition tasks~\cite{cummins_image-based_2017}. The neuron activations from a particular layer constitute the DS representation. For all of our DS representations we used the VGG19 model and output layer \textit{fc2}, with 4096-dimensions. This particular layer was the default setting, and represents the second fully-connected layer (near the output layer) which comes after the deep convolutional layers\footnote{https://github.com/DeepSpectrum/DeepSpectrum}. This particular image-based representation has been proven useful for detecting sleep apneas and snoring~\cite{amiriparian_snore_2017}, essentially modeling very fine-grained types of vibrations at a high resolution. We thought the DS features may also capture nuanced variations in synthetic speech as well. This type of feature vector is \textit{4096 x 1} dimensions.

\subsubsection{X-Vector Embeddings}\label{sec:xvector_descr}
In~\cite{williams2019speech} they proposed to extract \textit{x}-vectors that represent different environment and attack types. Building on this idea, we used the categorical labels of the PA training set as labels to train discriminative \textit{x}-vector models. We train six \textit{x}-vector extractors on the PA training set - we use the same architecture as in \cite{snyder_x-vectors:_2018}. The models differ by the objective to differentiate between:
\begin{itemize}
    \item speakers (xvec0),
    \item room size (xvec1),
    \item T60 reverberation time (xvec2),
    \item talker-to-ASV distance (xvec3),
    \item attacker-to-talker distance (xvec4),
    \item replay device quality (xvec5).
\end{itemize}
By training different \textit{x}-vector embeddings we aim to capture the representations that encapsulate the differentiating features for the above characteristics. By comparing different \textit{x}-vector types as the utterance embeddings for the task of automatic quality estimation, we can better understand the desired aspects of the optimal embeddings for this task. Each utterance-level \textit{x}-vector is \textit{512 x 1} dimensions.

\subsubsection{Acoustic Model (AM) Embeddings}
AM embeddings were proposed in~\cite{rownicka_analyzing_2018} as utterance embeddings for AM adaptation in ASR, and further explored in~\cite{rownicka2019embeddings} for the same task. They are extracted from five selected layers $i = \{3,6,9,12,15\}$ of a deep CNN acoustic model trained with cross-entropy (CE) criterion. To extract the utterance-level AM embedding, the representations at different layers $i$ are merged and reduced to 512 dimensions with a PCA transform. We trained on the AMI multiparty meetings corpus~\cite{ami_corpus}. To extract the AM embeddings we fix the parameters and perform a forward pass for each utterance. Each utterance-level embedding is \textit{512 x 1} dimensions.


We use those embeddings for automatic quality estimation in this work, since the embeddings in their raw form (i.e. using an unsupervised PCA transformation) are not extracted with the constraints to capture any specific characteristics, unlike the \textit{x}-vectors. The label supervision is not used in the AM embedding extraction. Therefore, they are potentially good generic utterance summaries, without imposing any specific characteristics. 

We hypothesize that AM embeddings may be well-suited for automatic quality estimation, especially because it is still not clear how best to represent speech for this task. We base our hypothesis on findings in~\cite{rownicka_analyzing_2018}, where the embeddings were evaluated in an acoustic condition verification task for the Aurora-4 dataset~\cite{aurora}. For 14 different acoustic conditions (different noise types and different microphones) the AM embeddings were better at differentiating between the acoustic condition classes than the i-vectors. We believe that this attribute of the embeddings can be useful for automatic quality estimation for TTS systems. On the other hand, in~\cite{rownicka_analyzing_2018} the embeddings were worse than the i-vectors in a speaker verification task. 

\section{Experiments}
The original MOSNet features and architectures were utilized, including the pre-trained VC models and re-trained models (on the LA dataset). In our experiments with new feature representations, we used a low-capacity CNN architecture. Our evaluation was based on the same metrics as reported in \cite{lo_mosnet:_2019}, plus additionally we included Kendall Tau to evaluate ordinal ranking. Our goal in these experiments was to find a representation that could perform best, especially as no such comparisons have been made before. Our final experiments include analyzing the MOS predictions to characterize speakers, and we reflect on how well this approach generalizes to our own multi-speaker TTS system and dataset in a separate, held-out speaker analysis task in Section~\ref{sec:speaker_analysis}. 

\subsection{Neural Network Architectures}

\subsubsection{Original MOSNet}
We used the original MOSNet neural network architecture described in detail in \cite{lo_mosnet:_2019}, as well as the corresponding code\footnote{https://github.com/lochenchou/MOSNet}. Three different architectures were used: BLSTM, CNN, and CNN-BLSTM combination. Mainly, we used this in two different ways. First, we employed the pre-trained models that came with the code, which had been developed for MOS on the Voice Conversion Challenge 2018. This included one pre-trained model for each architecture. Secondly, we experimented with re-training the entire network using the LA dataset. During this set of experiments, we used the original features and experimented only with batch size and the alpha ($\alpha$) parameter from the loss function. For $\alpha$, we tested values of: $[0.0, 0.5, 1.0]$, noting that the default from \cite{lo_mosnet:_2019} was $1.0$. This parameter is a weighting of the loss at the frame-level compared to the loss at the utterance level. A value of $1.0$ suggests that all frames in an utterance are equal.

\subsubsection{Low-Capacity CNN}
All of our other experiments, in which we compare 8 different representations of the speech, were based on a simple CNN architecture implemented with the Keras library~\cite{chollet2015keras} with TensorFlow backend~\cite{abadi2016tensorflow}. The CNN consisted of the following layers: Conv1D $\rightarrow$ Conv1D $\rightarrow$ MaxPooling $\rightarrow$ Conv1D $\rightarrow$ Conv1D $\rightarrow$ GlobalAveragePoolng $\rightarrow$ Dropout $\rightarrow$ Dense. 

We experimented with an optional batch normalization layer at the input. The MaxPooling pool size was set to 3. Each Conv1D layer used a kernel size of 10, with L2 regularization~\cite{ng2004feature} and ReLU activation~\cite{hahnloser2000digital}. We used Adam optimizer~\cite{adam} with learning rate $lr=0.0001$, and early-stopping monitored by validation loss (Mean Square Error). The parameters that we tested were: batch normalization (on/off), L2 value [$0.0001, 0.001, 0.01, 0.1$], dropout rate [$0.1, 0.2, 0.3$], number of filters [$16, 32, 64, 128$], and batch size [$16, 64, 128$],

\subsection{Metrics}
We report four metrics when discussing the results of our experiments. Three of these metrics are also reported for VC-based MOSNet in \cite{lo_mosnet:_2019}, so we can have a meaningful comparison especially because some of our experiments involve the MOSNet architecture. We also include a metric to evaluate rankings. \\
\textbf{Linear Correlation Coefficient (LCC)}
Also known as the Pearson correlation (\textit{r}), this metric assumes that there is a linear relationship between variables which may or may not be accurate. The values range between $-1$ and $+1$, where a score of $+1$ would indicate a perfect correlation. However, this metric alone can sometimes give a high score even if there is not actually a strong correlation.\\
\textbf{Spearman Rank Correlation Coefficient (SRCC)}
The Spearman rank correlation ($\rho$) metric is monotonic which makes it potentially more useful than LCC. Similar to LCC, the values range between $-1$ and $+1$, where a score of $+1$ would indicate a perfect correlation. It does not have a precise interpretation but it is non-parametric so could model different kinds of relationships. For example, if the distribution of MOS scores in poor-quality TTS is focused around particular values, but there is more diversity in high-quality TTS systems. \\
\textbf{Mean Square Error (MSE)}
For this metric, we simply measure the error between the mean predicted MOS score and the mean true MOS score. As a standalone metric, the MSE is not ideal for this task because it fails to capture information about the distribution of scores. For example, if true and predicted scores tend toward a central distribution, a good (low) MSE score may miss characterizing high and low outliers.\\
\textbf{Kendall Tau Rank Correlation (KTAU)}
We also report Kendall tau ($\tau$) for the purpose of evaluating rankings. Specifically, we use tau-b \cite{knight1966computer}. The Kendall tau score tends to be more robust than Spearman $\rho$ in terms of error sensitivity~\cite{croux2010influence}. It is known to be useful for ordinal rankings. To calculate the Kendall tau values, we first calculate the mean predicted and mean true MOS score for each speaker or system (we report both). We then sort them to obtain true rankings and predicted rankings. The Kendall tau is then computed based on these ordinal rankings. 

\subsection{Best Model Selection}
Given that there are several metrics reported, as well as different ways to aggregate results, we made some decisions about how to select our best performing model for each representation in our experiments. Of the metrics that were reported in previous work \cite{lo_mosnet:_2019,fu_quality-net:_2018,patton_automos:_2016} we believe SRCC does best at capturing information about rank and distribution, although we recognize that no metric is perfect. When selecting our best model configuration for each representation in the experiments, we based our decision on SRCC aggregated at the speaker-level because our end goal is to understand how different speakers contribute to the overall quality of our multi-speaker TTS system, rather than compare systems. At the same time, we do report the metrics aggregated at the system-level even though these did not influence our model selection. This allows us to compare to previous work, which did not report results for speakers. 

\subsubsection{System-Level vs. Speaker-Level Aggregation}
One of the reasons for aggregating predictions at the system-level is to characterize an entire system, and compare systems with one another. By aggregating scores at this level, the low-quality systems (such as \textbf{A08} HMM-based TTS) will stand out from the high-quality TTS systems (such as \textbf{A10} WaveRNN TTS). Thus the correlation metrics are measuring if the NN can characterize the average quality of a given system. A high correlation value at the system-level can be interpreted to mean that the NN effectively separates low-quality and high-quality systems. We expect that a high correlation score also means that our NN model can generalize well to new systems. 

When we aggregate scores at the speaker-level, this allows us to see how particular speakers perform in a particular system, or more generally across multiple systems. We might observe, for example, that some speakers always generate higher quality speech. In that case, we could make some inferences about how that speaker might do in a new TTS system - which is one of our goals for this work. Similarly, we may identify particular speakers who always produce low-quality speech in any TTS system, so that we could avoid those speakers in the future.

\subsection{Results}
The results from our experiments are shown in Table~\ref{tab:main_res}. 
For each feature representation, the best model and parameters were selected based on highest speaker-Level SRCC score (yellow). We also highlight the system-level SRCC score (pink). We identified the overall best representation as xvec5 (blue) because of the high SRCC scores at both system-level and speaker-level, as well as the very high Kendall tau score for ranking speakers.

\begin{table*}[hbt!]
\begin{tabular}{l|l|cccccccccc|}
\cline{2-12}
                                                               &                                       & \multicolumn{10}{c|}{\textbf{Representations}}                                                                                                                                                                                                                                                                                                                                                                                                                                                                                                                                                                         \\
                                                               & \multirow{-2}{*}{\textbf{Metric}}     & \begin{tabular}[c]{@{}c@{}}Pre-trained \\ (VC-CNN)\end{tabular}   & \begin{tabular}[c]{@{}c@{}}Re-trained \\ (LA-CNN)\end{tabular}  & xvec0                                                      & xvec1                                                      & xvec2                                                      & xvec3                                                      & xvec4                                                      & \cellcolor[HTML]{DAE8FC}\textbf{xvec5}                      & DS                                                          & AM                                     \\ \hline
\multicolumn{1}{|c|}{}                                         & \textbf{LCC}                          & \multicolumn{1}{c|}{0.122}                                   & \multicolumn{1}{c|}{0.717}                                  & \multicolumn{1}{c|}{0.537}                                  & \multicolumn{1}{c|}{0.751}                                  & \multicolumn{1}{c|}{0.495}                                  & \multicolumn{1}{c|}{0.532}                                  & \multicolumn{1}{c|}{0.820}                                  & \multicolumn{1}{c|}{\cellcolor[HTML]{DAE8FC}\textbf{0.776}}  & \multicolumn{1}{c|}{0.918}                                  & 0.788                                  \\
\multicolumn{1}{|c|}{}                                         & \cellcolor[HTML]{FFCCC9}\textbf{SRCC} & \multicolumn{1}{c|}{\cellcolor[HTML]{FFCCC9}\textbf{-0.027}} & \multicolumn{1}{c|}{\cellcolor[HTML]{FFCCC9}\textbf{0.868}} & \multicolumn{1}{c|}{\cellcolor[HTML]{FFCCC9}\textbf{0.659}} & \multicolumn{1}{c|}{\cellcolor[HTML]{FFCCC9}\textbf{0.725}} & \multicolumn{1}{c|}{\cellcolor[HTML]{FFCCC9}\textbf{0.368}} & \multicolumn{1}{c|}{\cellcolor[HTML]{FFCCC9}\textbf{0.412}} & \multicolumn{1}{c|}{\cellcolor[HTML]{FFCCC9}\textbf{0.868}} & \multicolumn{1}{c|}{\cellcolor[HTML]{DAE8FC}\textbf{0.709}}  & \multicolumn{1}{c|}{\cellcolor[HTML]{FFCCC9}\textbf{0.885}} & \cellcolor[HTML]{FFCCC9}\textbf{0.808} \\
\multicolumn{1}{|c|}{}                                         & \textbf{MSE}                          & \multicolumn{1}{c|}{1.513}                                   & \multicolumn{1}{c|}{1.277}                                  & \multicolumn{1}{c|}{1.437}                                  & \multicolumn{1}{c|}{1.003}                                  & \multicolumn{1}{c|}{1.208}                                  & \multicolumn{1}{c|}{1.453}                                  & \multicolumn{1}{c|}{0.968}                                  & \multicolumn{1}{c|}{\cellcolor[HTML]{DAE8FC}\textbf{0.859}}  & \multicolumn{1}{c|}{0.388}                                  & 0.799                                  \\
\multicolumn{1}{|c|}{\multirow{-4}{*}{\textbf{System-Level}}}  & \textbf{KTAU}                       & \multicolumn{1}{c|}{-0.052}                                  & \multicolumn{1}{c|}{-0.026}                                 & \multicolumn{1}{c|}{-0.182}                                 & \multicolumn{1}{c|}{-0.026}                                 & \multicolumn{1}{c|}{0.130}                                  & \multicolumn{1}{c|}{0.000}                                  & \multicolumn{1}{c|}{0.000}                                  & \multicolumn{1}{c|}{\cellcolor[HTML]{DAE8FC}\textbf{-0.156}} & \multicolumn{1}{c|}{0.091}                                  & -0.208                                 \\ \hline
\multicolumn{1}{|l|}{}                                         & \textbf{LCC}                          & \multicolumn{1}{c|}{0.006}                                   & \multicolumn{1}{c|}{0.815}                                  & \multicolumn{1}{c|}{0.465}                                  & \multicolumn{1}{c|}{0.011}                                  & \multicolumn{1}{c|}{0.426}                                  & \multicolumn{1}{c|}{0.267}                                  & \multicolumn{1}{c|}{0.474}                                  & \multicolumn{1}{c|}{\cellcolor[HTML]{DAE8FC}\textbf{0.694}}  & \multicolumn{1}{c|}{0.395}                                  & 0.398                                  \\
\multicolumn{1}{|l|}{}                                         & \cellcolor[HTML]{FFFFC7}\textbf{SRCC} & \multicolumn{1}{c|}{\cellcolor[HTML]{FFFFC7}\textbf{-0.033}} & \multicolumn{1}{c|}{\cellcolor[HTML]{FFFFC7}\textbf{0.883}} & \multicolumn{1}{c|}{\cellcolor[HTML]{FFFFC7}\textbf{0.433}} & \multicolumn{1}{c|}{\cellcolor[HTML]{FFFFC7}\textbf{0.017}} & \multicolumn{1}{c|}{\cellcolor[HTML]{FFFFC7}\textbf{0.383}} & \multicolumn{1}{c|}{\cellcolor[HTML]{FFFFC7}\textbf{0.367}} & \multicolumn{1}{c|}{\cellcolor[HTML]{FFFFC7}\textbf{0.717}} & \multicolumn{1}{c|}{\cellcolor[HTML]{DAE8FC}\textbf{0.800}}  & \multicolumn{1}{c|}{\cellcolor[HTML]{FFFFC7}\textbf{0.500}} & \cellcolor[HTML]{FFFFC7}\textbf{0.583} \\
\multicolumn{1}{|l|}{}                                         & \textbf{MSE}                          & \multicolumn{1}{c|}{0.126}                                   & \multicolumn{1}{c|}{0.053}                                  & \multicolumn{1}{c|}{0.111}                                  & \multicolumn{1}{c|}{0.194}                                  & \multicolumn{1}{c|}{0.074}                                  & \multicolumn{1}{c|}{0.095}                                  & \multicolumn{1}{c|}{0.185}                                  & \multicolumn{1}{c|}{\cellcolor[HTML]{DAE8FC}\textbf{0.091}}  & \multicolumn{1}{c|}{0.126}                                  & 0.104                                  \\
\multicolumn{1}{|l|}{\multirow{-4}{*}{\textbf{Speaker-Level}}} & \textbf{KTAU}                       & \multicolumn{1}{c|}{-0.057}                                  & \multicolumn{1}{c|}{0.114}                                  & \multicolumn{1}{c|}{-0.400}                                 & \multicolumn{1}{c|}{0.514}                                  & \multicolumn{1}{c|}{0.057}                                  & \multicolumn{1}{c|}{0.114}                                  & \multicolumn{1}{c|}{0.343}                                  & \multicolumn{1}{c|}{\cellcolor[HTML]{DAE8FC}\textbf{0.771}}  & \multicolumn{1}{c|}{-0.057}                                 & -0.114                                 \\ \hline
\end{tabular}
\caption{\label{tab:main_res}System-level and speaker-level prediction results on the LA dataset, comparing original pre-trained VC-CNN model, re-trained LA-CNN model, and 8 new representations from best parameters on our CNN architecture. For each, the best model and parameters were selected based on highest speaker-level SRCC score (yellow). }
\end{table*}

Overall, many of the representations demonstrate similar performance for our chosen metrics. For example, at first glance there appears to be only minor differences between re-trained LA-CNN and the DS representation. As a regression task, the DS representation yields a lower MSE which may suggest that the predicted MOS values are closer to target. The DS representation appears to perform well at the system-level, however it is not good at the speaker-level. Therefore several representations stand out that are clearly not suited for distinguishing quality for either systems or speakers: pre-trained VC-CNN, xvec0, xvec2 and xvec3.

The xvec5 representation (highlighted blue in Table~\ref{tab:main_res}) provided the highest Kendall tau score for speakers. We further examined the ranking distributions for individual speakers per-system. From the ground-truth MOS scores, we focused on the following two systems:
\begin{itemize}

\item \textbf{A08}: poorest quality system, HMM-based TTS ($MOS_{mean}=1.79$)
\item \textbf{A10}: highest quality system, WaveRNN TTS ($MOS_{mean}=5.58$) 
\end{itemize}

\begin{figure}[h]
    \centering
    \includegraphics[width=1\linewidth]{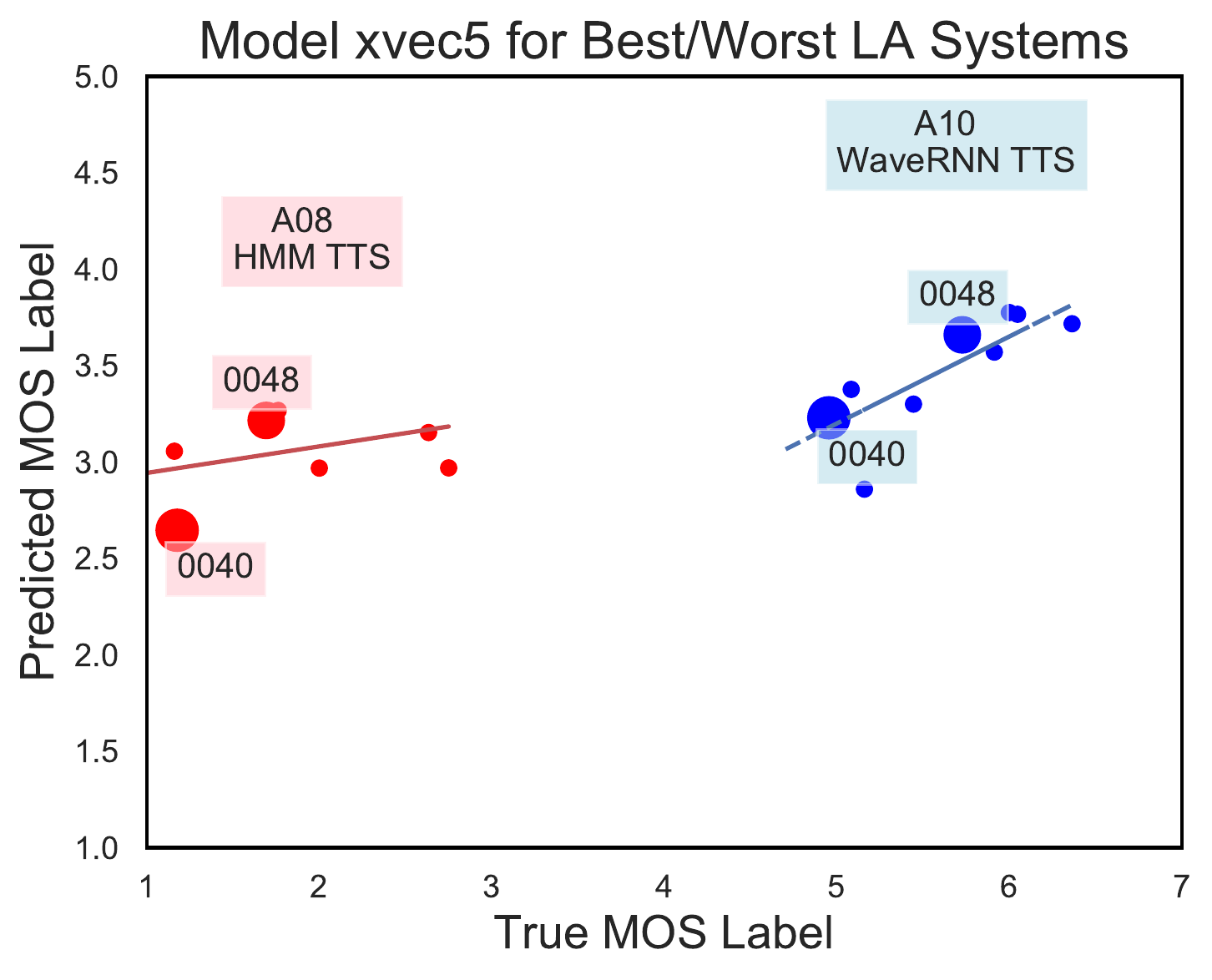}
    \caption{Comparison of two systems from the LA dataset. The selection of best and worst was based on average ground-truth MOS scores. The system \textbf{A10} (WaveRNN TTS) was rated by humans as the best system in the LA dataset, while \textbf{A08} (HMM-based TTS) was rated as the worst.}
    \label{fig:LA_systems}
\end{figure}

From Figure~\ref{fig:LA_systems} we see the individual speakers for each of the best and worst LA systems. Speaker \textit{0048} has the highest overall MOS ratings across all systems in the LA dataset, while speaker \textit{0040} has the lowest. We observe from Figure~\ref{fig:LA_systems} that the relative relationship between speaker qualities is preserved. Speaker \textit{0048} is always predicted to perform better than speaker \textit{0040} on average. Of course this difference is relative within a particular TTS system. The worst speakers in \textbf{A10} are as good as, or better, than the best speakers in \textbf{A08}. Since these relationships between speakers are relative to a given TTS system, it would not be possible to generalize across all speakers and all systems in the LA dataset - that is why we observe a negative Kendall tau score of $-0.156$ for the system-level xvec5 representation. We report that the best-performing model (xvec5) used a batch size of $1$, with $16$ filters, dropout rate of $0.2$ and L2 value of $0.0001$, without batch normalization. Likewise for the reported re-trained LA-CNN model, the best performance used a CNN and $\alpha$ = $1.0$ for equal frame weighting (as in the original MOSNet paper). 

Another important note is that the average MOS scores for speakers within a given system are not particularly spread out. Consider that the range of true MOS scores for \textbf{A10} is similar to the range of true MOS scores for \textbf{A08}. While this is very helpful for characterizing systems, it does little for characterizing particular speakers. Therefore we recommend that future discussions about speaker characterization be kept within the context of a particular TTS system. 

In the LA dataset, the average system based on median MOS score was \textbf{A11} with $MOS_{mean}=2.41$ (encoder-decoder sequence modeling TTS with attention). And the average system based on mean MOS score was \textbf{A07} with $MOS_{mean}=3.65$ (Merlin DNN SPSS with speaker codes and WORLD vocoder).

\section{Speaker Analysis}\label{sec:speaker_analysis}
Our experiments with different representations have so far shown that we can use a NN to predict MOS scores such that we can differentiate one system from another. In this section we further expand our analysis to characterize speakers, as well as to explore how well our MOS models can generalize to our own multi-speaker TTS system.

\subsection{Characterizing Speakers from LA dataset}\label{sec:la_spk_characterization}
\begin{figure}[h]
    \centering
    \includegraphics[width=1\linewidth]{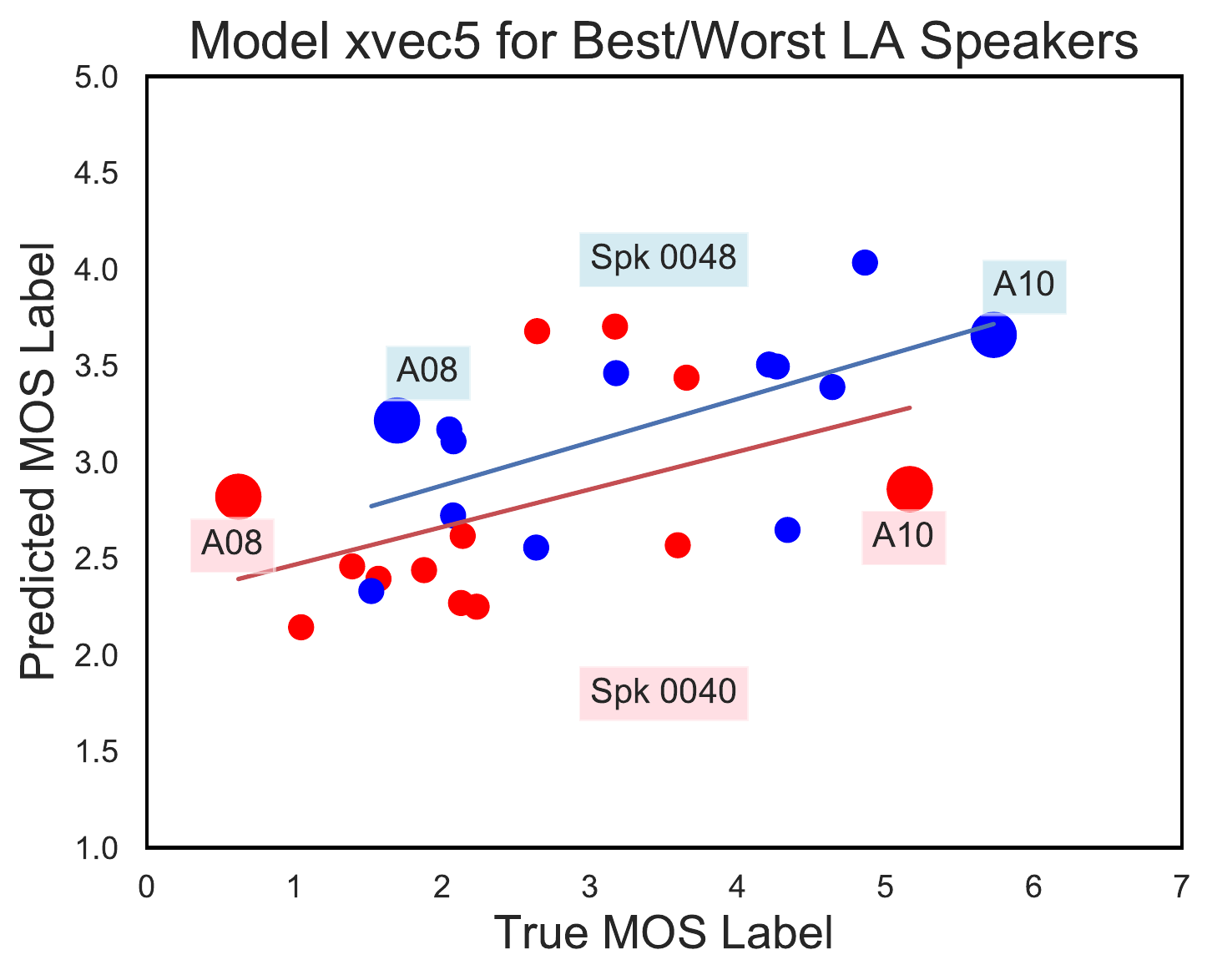}
    \caption{Comparison of two speakers from the LA dataset. The selection of best (\textit{0048}) and worst (\textit{0040}) speakers was based on average ground-truth MOS scores. We also note each speaker's performance on the best (\textbf{A10}) and worst (\textbf{A08}) TTS systems.}
    \label{fig:LA_speakers}
\end{figure}

Our NN correctly predicts that speakers \textit{0046}, \textit{0047}, and \textit{0048} will tend to have higher MOS scores than other speakers, especially for neural TTS systems (\textbf{A12}, \textbf{A15}, \textbf{A10}). Our system also correctly predicts that speakers \textit{0040}, \textit{0042}, and \textit{0044} will tend to have lower MOS scores compared to the other speakers. We observe this consistently when we use our NN trained with the \textbf{xvec5} representation, and demonstrate this finding in Figure~\ref{fig:LA_speakers}. This plot shows that speaker \textit{0040} true and predicted MOS scores are generally lower than those same scores for speaker \textit{0048}. Therefore we could suggest that speaker \textit{0048} is contributing to a performance increase for most systems - for ground-truth and predicted MOS. 

\begin{table*}[h]
\begin{tabular}{c|l|cccccccccc|}
\cline{2-12}
\multicolumn{1}{l|}{}                                          & \multicolumn{1}{c|}{}                                  & \multicolumn{10}{c|}{\textbf{Representations}}                                                                                                                                                                                                                                                                                                                                          \\
\multicolumn{1}{l|}{}                                          & \multicolumn{1}{c|}{\multirow{-2}{*}{\textbf{Metric}}} & \begin{tabular}[c]{@{}c@{}}Pre-trained\\ (VC-CNN)\end{tabular} & \begin{tabular}[c]{@{}c@{}}Re-trained\\ (LA-CNN)\end{tabular} & xvec0                     & xvec1                      & xvec2                      & xvec3                      & xvec4                      & \cellcolor[HTML]{DAE8FC}\textbf{xvec5}                     & DS                          & AM    \\ \hline
\multicolumn{1}{|c|}{}                                         & \textbf{LCC}                                           & \multicolumn{1}{c|}{0.570}                                     & \multicolumn{1}{c|}{0.104}                                    & \multicolumn{1}{c|}{0.525} & \multicolumn{1}{c|}{0.439}  & \multicolumn{1}{c|}{0.155}  & \multicolumn{1}{c|}{-0.021} & \multicolumn{1}{c|}{-0.322} & \multicolumn{1}{c|}{\cellcolor[HTML]{DAE8FC}\textbf{0.096}} & \multicolumn{1}{c|}{0.319}  & 0.603 \\
\multicolumn{1}{|c|}{}                                         & \textbf{SRCC}                                          & \multicolumn{1}{c|}{0.534}                                     & \multicolumn{1}{c|}{0.102}                                    & \multicolumn{1}{c|}{0.630} & \multicolumn{1}{c|}{0.371}  & \multicolumn{1}{c|}{-0.042} & \multicolumn{1}{c|}{-0.060} & \multicolumn{1}{c|}{-0.392} & \multicolumn{1}{c|}{\cellcolor[HTML]{DAE8FC}\textbf{0.074}} & \multicolumn{1}{c|}{0.347}  & 0.632 \\
\multicolumn{1}{|c|}{}                                         & \textbf{MSE}                                           & \multicolumn{1}{c|}{4.144}                                     & \multicolumn{1}{c|}{3.343}                                    & \multicolumn{1}{c|}{3.503} & \multicolumn{1}{c|}{3.435}  & \multicolumn{1}{c|}{4.879}  & \multicolumn{1}{c|}{3.937}  & \multicolumn{1}{c|}{3.425}  & \multicolumn{1}{c|}{\cellcolor[HTML]{DAE8FC}\textbf{4.320}} & \multicolumn{1}{c|}{6.354}  & 6.240 \\
\multicolumn{1}{|c|}{\multirow{-4}{*}{\textbf{Speaker-Level}}} & \textbf{KTAU}                                        & \multicolumn{1}{c|}{-0.138}                                    & \multicolumn{1}{c|}{-0.032}                                   & \multicolumn{1}{c|}{0.190} & \multicolumn{1}{c|}{-0.063} & \multicolumn{1}{c|}{-0.138} & \multicolumn{1}{c|}{-0.106} & \multicolumn{1}{c|}{-0.106} & \multicolumn{1}{c|}{\cellcolor[HTML]{DAE8FC}\textbf{0.296}} & \multicolumn{1}{c|}{-0.159} & 0.032 \\ \hline
\end{tabular}
\caption{\label{tab:ophelia_res}Speaker-level prediction results on Ophelia/LibriTTS synthetic speech. The performance shown here is based on models that were originally trained on the LA dataset (as reported in Table~\ref{tab:main_res}) before being used to evaluate our multi-speaker TTS from Ophelia/LibriTTS.}
\end{table*}

\subsection{Multi-Speaker Seq-to-Seq TTS}
\label{sec:ophelia_description}
We trained a TTS system using the Deep Convolutional TTS (DCTTS) \cite{tachibana2018efficiently}, implemented in Ophelia \cite{ophelia}, a sequence-to-sequence with attention model. We trained a multi-speaker system on a subset of 60 hrs from the LibriTTS corpus \cite{zen2019libritts}, made of 37000 utterances and 145 speakers. Ophelia/DCTTS corresponds to a state-of-the-art, fast trainable TTS system, and therefore a good choice among the current architectures \cite{Watts2019}. As mentioned in section \ref{sec:libri}, the diversity present in the LibriTTS dataset makes it a good candidate to test our automatic evaluation system. Our overall goal has been to automatically evaluate the quality of our own multi-speaker TTS system. The relevance of running such evaluation automatically stems from the fact that our system can generate 145 speakers, however, it would be quite costly to evaluate multiple samples from each one of them into a full standard listening test. An evaluation metric would give us an idea of the best and worst speakers for our system.

\subsection{Our MOS test}
Ideally, our NN would be able to predict MOS scores so reliably that we would not need to conduct listening tests in our lab on a regular basis. We needed to examine whether or not our models could be used on our own TTS system. We conducted a small MOS test to verify this. We attempted to replicate aspects of the MOS tests that were performed when rating the LA dataset. For example, we used the same scale of 1-10 for naturalness. We advised our listeners that some speech samples could be from a machine or a human, however our listeners were only presented with TTS speech. The instructions were given as: \textit{``Rate the following speech samples based on whether or not they sound more like a human (10) or a machine (1). Consider that some samples may sound like a human but have a poor quality of recording or other noise.''}. These instructions were slightly different than the instructions used for rating the LA dataset due to the content and meaning of the Harvard Sentences, as it would not appropriately fit into a ``call center'' type of scenario. 

The test materials consisted of 20 TTS speakers who were selected randomly from the 145 used for training. For each speaker, we synthesized the first 100 Harvard Sentences\footnote{https://www.cs.columbia.edu/~hgs/audio/harvard.html}. We recruited 20 human listeners, aged 18 or over, from the University of Edinburgh community. All of our human listeners self-identified as a native-speaker, or near-native speaker of English. Each human subject provided a MOS score (on a scale of 1-10), for 100 utterances. These were balanced across the 20 synthetic speakers, and randomized. Therefore, in line with the LA dataset ratings, each utterance was marked only once. To present the MOS test to our human listeners, we used a simple web-based interface
from \cite{schoeffler2018webmushra}

\subsection{Characterizing Speakers from the LibriTTS Dataset}
\begin{figure}[h]
    \centering
    \includegraphics[width=1\linewidth]{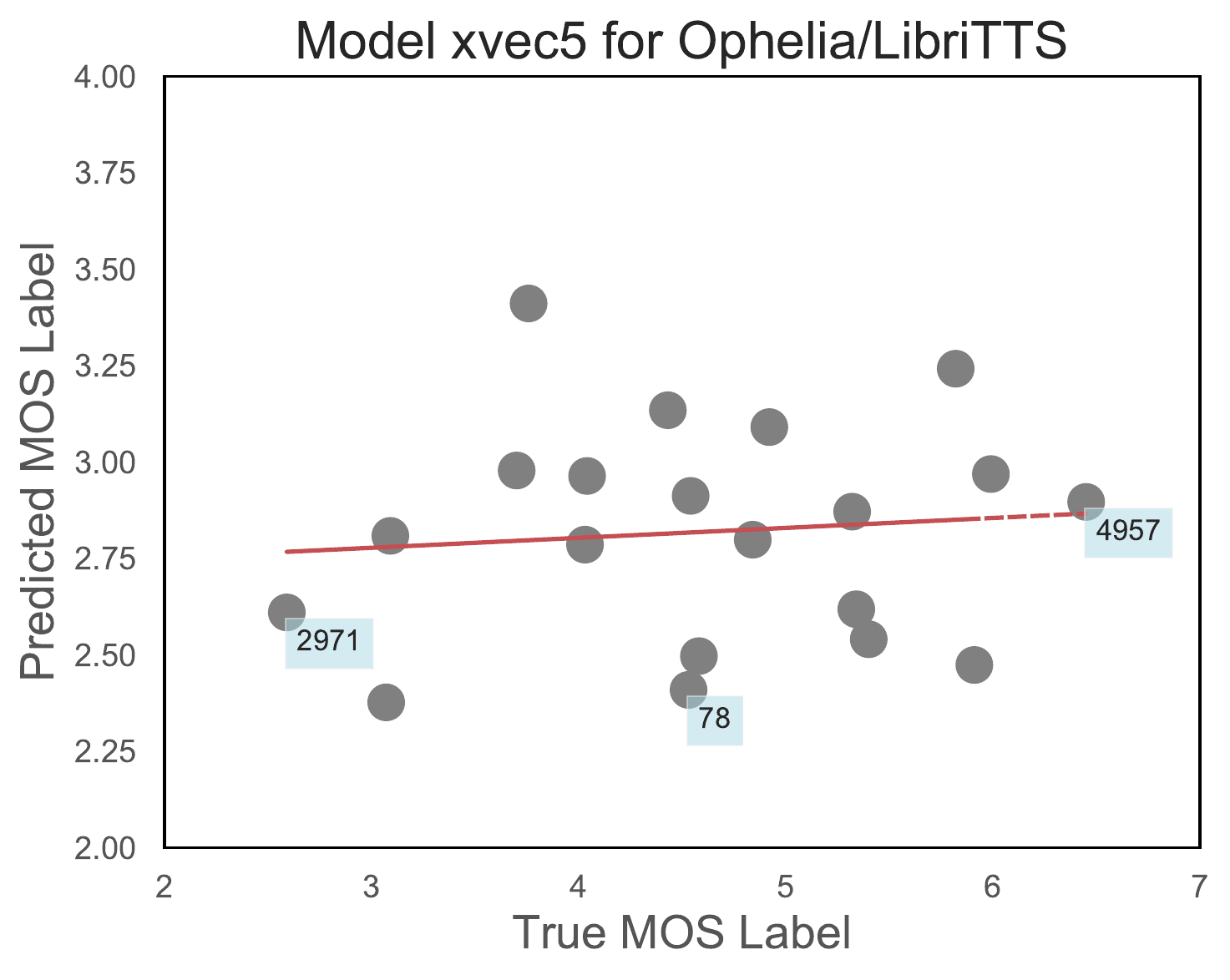}
    \caption{Spread of speakers for Ophelia/LibriTTS. The speakers are marked on the plot for those MOS scores of best (speaker \textit{4957}), worst (speaker \textit{2971}), and mean (speaker \textit{78}). Note that we have adjusted the axis bounds for readability purposes.} 
    \label{fig:ophelia_speakers}
\end{figure}

Similar to our earlier analysis of speakers in Section~\ref{sec:la_spk_characterization}, we visualized our MOS predictions in Figure~\ref{fig:ophelia_speakers}. In this figure, we have highlighted the best (\textit{4957}), worst (\textit{2971)} and average/mean (\textit{78}) as determined from the human judgments. Most obvious is that while the human judgments reflect a spread over a range of MOS scores, our best NN model still only predicted MOS scores within a central distribution. This indicates that our approach for predicting MOS is not reliable. However the model can learn to rank the relative speaker quality (i.e. speaker \textit{4957} is predicted to be higher quality than \textit{2971}).

\subsection{Summary}
\begin{figure}[h]
    \centering
    \includegraphics[width=1\linewidth]{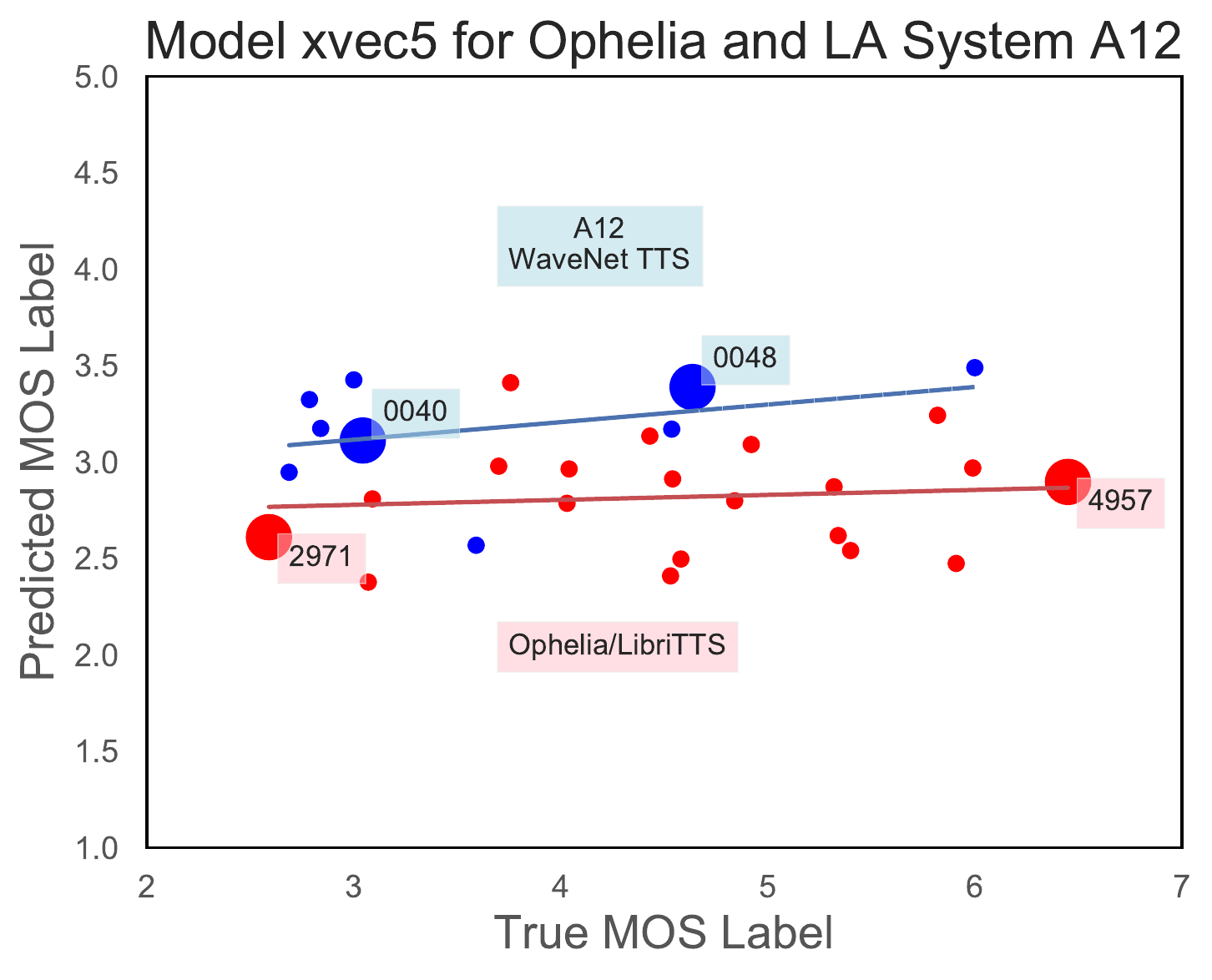}
    \caption{WaveNet and Ophelia side-by-side - our TTS system can produce some speakers with high quality comparable to WaveNet TTS. In both cases it is difficult to automatically rate the quality given this very small positive correlation. Selectively synthesizing particular speakers could artificially boost the overall MOS score for any system.} 
    \label{fig:ophelia_A12}
\end{figure}

Speaker analysis, which had not been done previously, has allowed us to examine the MOS prediction task from a new angle. The ability to rank speakers is potentially very useful for numerous development and evaluation tasks. Another finding that caught our attention in Table~\ref{tab:ophelia_res} is that the pre-trained VC-CNN MOSNet model appears to perform better on our Ophelia TTS system, than our re-trained LA-CNN - at least according to the LCC and SRCC metrics. We identified two TTS systems from the LA dataset which the pre-trained VC-CNN model performed best, and not surprisingly one of them is a similar architecture to our Ophelia TTS (\textbf{A11}: encoder-decoder using sequence-to-sequence modeling with attention). We identified system \textbf{A12} from the LA dataset which had the most similar performance to our Opehlia/LibriTTS. We further observed that for both of these TTS systems, our NN could rank best/worst speakers, though the ability to rank could be greatly improved for both. This relationship is visualized in Figure~\ref{fig:ophelia_A12}. Also importantly, we noted that for the true MOS labels, our Ophelia/TTS system could produce speakers who are similar quality to the WaveNetTTS speakers. Of course, these two systems underwent different MOS tests at different times, therefore we believe a new approach to scoring is a needed research direction. 

\section{Conclusions and Future Work}
This paper has described our work and analysis towards achieving a method that we can use to reliably rate our TTS quality automatically. Having such a capability would save resources, such as time and costs, and would otherwise facilitate rapid analysis and evaluation. It is important to note that based on our analysis of re-training the MOSNet architecture on the LA dataset, we find that the technique overall is sound. However we could not train the models to generalize well to a new dataset and system, as we demonstrated with our Ophelia/LibriTTS experiments. From this outcome, we might suggest including a pre-trained MOSNet model, using a relevant dataset when releasing a new TTS system, or to simply re-train a MOSNet when switching from one dataset to the next.

Further, the research direction of relative quality rankings could be useful. Instead of predicting a MOS score during a regression task, one could instead predict relative rankings akin to A/B testing. This would potentially require a different set of metrics and a different loss function as well. Another related problem that this could be used for, is to pre-select speakers for TTS training based on the quality of their recordings. While the LibriTTS dataset underwent some cleaning~\cite{zen2019libritts}, it is widely accepted that some speakers in the dataset are better-suited for TTS than others, and we need a way to identify them. While have trained an automatic systems that can rank speakers according to the human perception in the MOS tests, we are not able to provide a tool that generalizes well to new TTS systems.

Finally, from our experiments on the LA dataset, we know that some embeddings seem to do better at characterizing TTS systems very well (DS) whereas others tend to characterize speakers (xvec5). We propose to further investigate the design of the optimal embeddings to represent utterances for automatic quality estimation. The results in this work have revealed some of the desired properties of such embeddings, e.g. \textit{x}-vectors for the device quality were superior to all other \textit{x}-vectors. Based on those findings, the other embeddings can be altered such that they capture more differentiating features w.r.t. the device quality. This could be done by applying an LDA transform, supervised by the device quality labels, to the embeddings. We leave those experiments as future work.

\section{Acknowledgments}
We thank Junichi Yamagishi and Xin Wang at National Institute for Informatics (NII) for providing MOS test scores on the ASVSpoof 2019 LA dataset, as well as for their helpful discussions. We also want to thank Cassia Valentini-Botinhao, Erfan Lowemi and Carol Chermaz at University of Edinburgh Centre for Speech Technology Research (CSTR) for their helpful discussions. This work was supported in part by the EPSRC Centre for Doctoral Training in Data Science, funded by the UK Engineering and Physical Sciences Research Council (grant EP/L016427/1) and the University of Edinburgh. It was also supported by a PhD studentship from the DataLab Innovation Centre, Ericsson Media Services, and Quorate Technology. This work was supported in part by CONICYT, Becas Chile, nº 72190135. 

\bibliographystyle{IEEEbib}
\bibliography{references}
\end{document}